\documentclass{article}
\usepackage{spconf,amsmath,graphicx}
\usepackage{color}
\usepackage{multirow}
\usepackage{floatrow}

\title{Explore the prower of dropout on Few-shot Learning}
\name{Shaobo Lin \qquad
	Xingyu Zeng \qquad
	Rui Zhao}
\address{Sensetime Research}
\begin{document}
	%
	\maketitle
	%

\begin{abstract}
	The generalization power of the pre-trained model is the key for few-shot deep learning. Dropout is a regularization technique used in traditional deep learning methods. In this paper, we explore the power of dropout on few-shot learning and provide some insights about how to use it. 
	 Extensive experiments on the few-shot object detection and few-shot image classification datasets, \emph{i.e.}, Pascal VOC, MS COCO, CUB, and mini-ImageNet, validate the effectiveness of our method.
\end{abstract}

\section{Introduction}
\label{submission}

Deep Neural Networks (DNNs) have achieved great progress in many computer vision tasks ~\cite{ren2016faster,dai2016r}. However, the impressive performance of these models largely relies on a large amount of data as well as expensive human annotation. When the annotated data are scarce, DNNs cannot generalize well to testing data especially when the testing data belongs to different classes of the training data. In contrast, humans can learn to recognize or detect a novel object quickly with only a few labeled examples. Due to some object categories naturally have few samples or their annotations are extremely hard to obtain, the generalization ability of conventional neural networks is far from satisfactory.

\begin{figure}[t]
	\begin{center}
		\centerline{\includegraphics[width=\columnwidth]{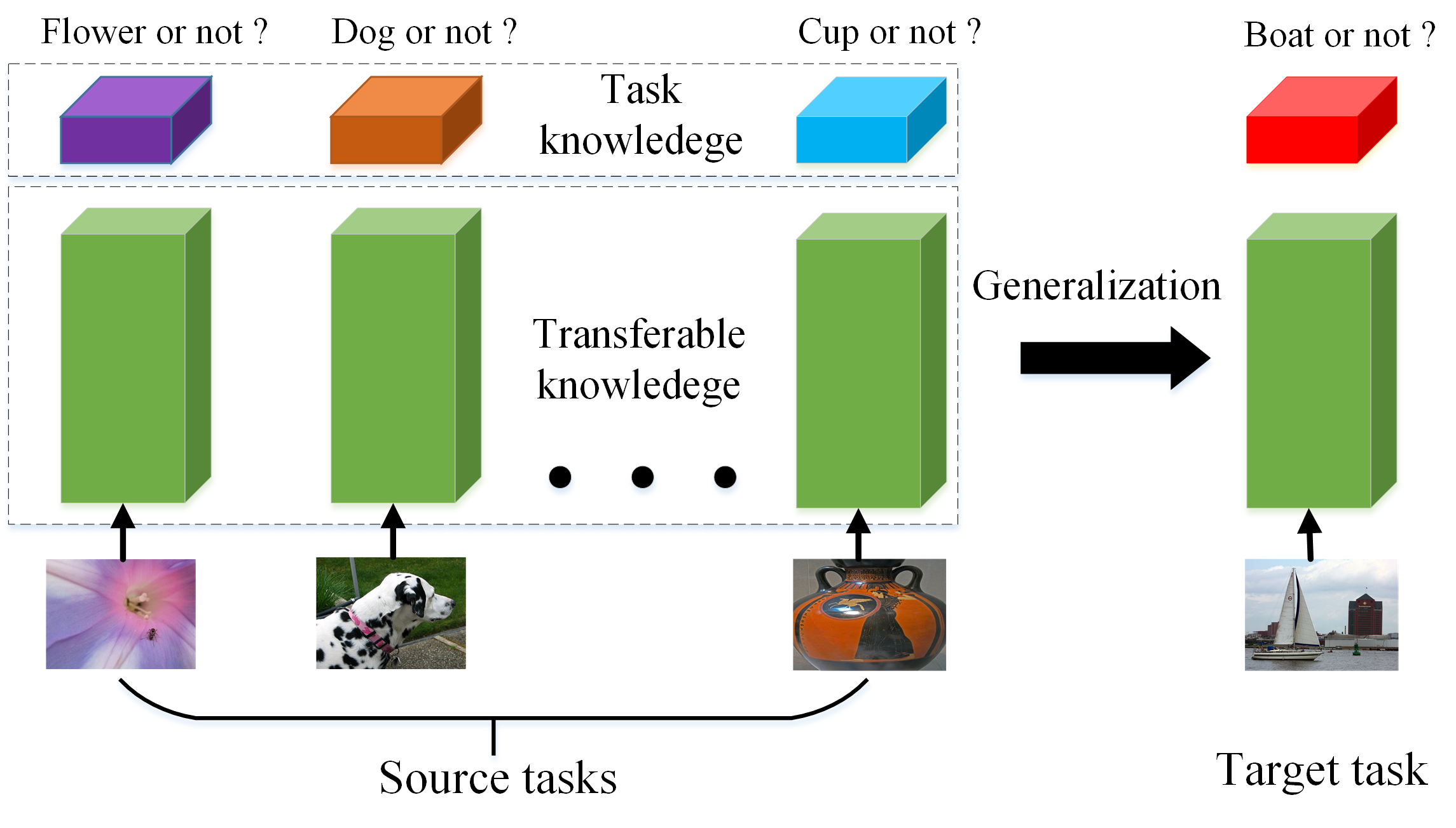}}
		\caption{The generalization power of transferable knowledge across different source tasks is the key for few-shot learning, in which transferable knowledge is adapted to the target task.}
		\label{fig1}
	\end{center}
\end{figure}

Few-shot learning, therefore, becomes an important research topic to achieve better generalization ability by learning from only a few examples. 
The mainstream few-shot learning approaches have two training stages: the pre-training stage and fine-tuning stage, the former is responsible for obtaining a good initialization point from data-sufficient base classes, and the latter adapts the pre-trained model to a specific task based on data-scarce novel classes. 
As shown in Fig.\ref{fig1}, a few-shot  model can be divided into two components, called transferable knowledge and task-specific knowledge. The former  represents the well-learned feature representation which is needed to be generalized to novel tasks in pre-training stage, the latter is the fine-tuned part of a model for specific categories during fine-tuning stage. Therefore, it is crucial to improve the generalization power of transferable knowledge if we want to apply it from source tasks to novel tasks in few-shot learning. We select several different few-shot methods as the baselines to validate the effectiveness of dropout, and explore how to use it in few-shot learning. By utilizing dropout, our model demonstrates great superiority towards the current excellent methods on few-shot object detection and image classification tasks. Dropout is not a new idea. However, we use it to solve a new problem (few-shot learning) and we provide more insights about how to use it, which are the major novelties. Our contributions can be summarized as three-fold. First, we introduce the idea of dropout to improve the generalization power of transferable knowledge in a new task: few-shot learning. Second, we provide some insights about how to use dropout in few-shot learning methods. For example, dropout is applied to pre-training stage, not fine-tuning stage.  Finally, experiments evaluate the effectiveness of our approach on the few-shot object detection and image classification datasets, \emph{i.e.}, Pascal VOC, MS COCO, CUB, and mini-ImageNet. 

\section{Method}

\subsection{The Solution of Few-shot Problems}
Common supervised learning problems based on abundant training data can be solved by minizing the Equ (1), in which D is a training dataset, $\Theta$ is the trainable parameters of neural networks, and x is input data sampled from $p(D)$. While few-shot tasks are limited in the number of samples, optimizing it directly is likely to results in the overfitting of $\Theta$ due to its high-dimensional property.

\begin{equation}
	\underset{\Theta}{min}\underset{x \sim p(D)}{E}{L(\Theta; x)}
\end{equation}   

Few-shot learning aims to solve learning problems with just a few training examples. To achieve this goal, the common solution is to reduce the learnable dimension of $\Theta$, and thus the Equ (1) can be re-written as Equ (2), in which $\Theta$ = [$\theta$,$w$]. 	 
$w$ represents the useful foundation for few-shot learning, such as a well-learned feature representation, which is obtained from source tasks. $\theta$ are updated on target tasks based on the learned $w$.

\begin{equation}
	\underset{\theta}{min}\underset{x \sim p(D)}{E}{L(\theta;x | w)}
\end{equation}



\subsection{Dropout on Few-shot Learning} 
Dropout~\cite{srivastava2014dropout} is applied to the transferable knowledge (w) which is trained with abundant source datasets. By applying dropout, we can obtain a new formulation of few-shot learning below.

\begin{equation}
	\underset{\theta}{min}\underset{x \sim p(D)}{E}{L(\theta;x | O(w))}
\end{equation}
$O$ is applying dropout on $w$. We call our dropout as w-dropout.

Dropout is not a novel idea for generalizing the model. However, the key question is how to use it to solve what we face to and why we can obtain the better performance. Dropout is used to solve new problems and we provide more insights in the following sections about how to use it. Moreover, the utilization of dropout in our work is different from normal dropout, since w-dropout is applied to transferable knowledge during pre-training stage, while normal dropout is commonly applied to task-specific knowledge in fine-tuning stage. By using w-dropout, the degree of over-fitting to base classes can be relieved, thereby the model is easier to be adapted to novel classes. That is why our w-dropout is better than normal dropout in few-shot models.

\section{Few-shot Object Detection}
\subsection{Datasets}
We evaluate our methods on Pascal VOC~\cite{everingham2010pascal} and MS COCO~\cite{lin2014microsoft}. 
In PASCAL VOC, we adopt the common strategy~\cite{ren2016faster,dai2016r} that using VOC 2007 test set for evaluating while VOC 2007 and 2012 train/val sets are used for training. Following~\cite{yan2019meta}, 5 out of its 20 object categories are selected as the novel classes, while keeping the remaining 15 ones as the base classes. We evaluate with three different novel/base splits from ~\cite{yan2019meta}, named as split 1, split 2, and split 3. 
Following~\cite{yan2019meta,wang2020frustratingly,sun2021fsce},  we use the mean average precision (mAP) at 0.5 IoU threshold as the evaluation metric, and report the results on the official test set of VOC 2007.
When using MS COCO, 20 out of 80 categories are reserved as novel classes, the rest 60 categories are base classes. The  detection performance with COCO-style AP, AP50, and AP75 for K = 10 and 30 shots of novel categories are reported.

\setlength{\tabcolsep}{3pt}
\begin{table}[t]
	\scriptsize
	
	\begin{center}
		\caption{Comparison with state-of-the-art few-shot object detection methods on VOC2007 test set for novel classes of the three splits. $^\Delta$ represents running on one selected seed, and others are averaged over 10 seeds.  {\color{red}{\bf Red}} and {\bf black} indicate state-of-the-art in the setting of single or multiple run. * means applying our w-dropout.}
		\label{table:1}
		\begin{tabular}{lccccccccc}
			\hline\noalign{\smallskip}
			&\multicolumn{3}{c}{split 1} & 	\multicolumn{3}{c}{split 2}&\multicolumn{3}{c}{split 3}\\
			\hline\noalign{\smallskip}
			Method/Shot& 1 & 3 & 10 & 1 &3&10&1&3&10\\
			\noalign{\smallskip}
			\hline
			\noalign{\smallskip}
			FR~\cite{kang2019few}&14.8&26.7&47.2&15.7&22.7&40.5&21.3&28.4&45.9\\
            FRCN+ft~\cite{kang2019few}&11.9&29&36.9&5.9&23.4&28.8&5.0&18.1&43.4\\
			Meta R-CNN~\cite{yan2019meta}&19.9&35&51.5&10.4&29.6&45.4&14.3&27.5&48.1\\
			Meta R-CNN (Our Impl.)&14.7&32.8&51.9&13.5&24.3&39.2&15.4&33.8&43.1\\
			Meta R-CNN* &24.7&37.3&51.8&16.8&28.9&40.1&{\bf20.2}&35.4&45.7\\
		    \hline
			\noalign{\smallskip}
			TFA~\cite{wang2020frustratingly}&25.3&42.1&52.8&{\bf18.3}&{\bf30.9}&39.5&17.9&34.3&45.6\\
			TFA (Our Impl.)&22.4&40.3&53.1&15.6&26.7&37.4&16.9&32.3&47.7\\
			TFA*&{\bf26.3}&{\bf45.6}&{\bf55.8}&15.9&29.9&{\bf40.8}&20.1&{\bf36.7}&{\bf49}\\
		    \hline
			\noalign{\smallskip}
			CME$^\Delta$~\cite{li2021beyond}&41.5&50.4&60.9&27.2&{\color{red}{\bf41.4}}&46.8&34.3&45.1&51.5\\
			MPSR$^\Delta$~\cite{wu2020multi}&41.7&{\color{red}{\bf51.4}}&61.8&24.4&39.2&47.8&35.6&42.3&49.7\\
			FSCN$^\Delta$~\cite{li2021few}&40.7&46.5&{\color{red}{\bf62.4}}&{\color{red}{\bf27.3}}&40.8&46.3&31.2&43.7&55.6\\	
			HallucFsDet$^\Delta$~\cite{zhang2021hallucination}&{\color{red}{\bf47}}&46.5&54.7&26.3&37.4&41.2&40.4&43.3&49.6\\
			FSCE$^\Delta$~\cite{sun2021fsce} (Our Impl.)&40.3&47.8&62.2&18.9&39.6&{\color{red}{\bf49.6}}&35.2&44.8&56.1\\				
			FSCE$^\Delta$*&44.6&50.3&61.7&24.6&40.8&49.4&{\color{red}{\bf41.1}}&{\color{red}{\bf46.2}}&{\color{red}{\bf57.1}}\\
			
			\hline	
		\end{tabular}
	\end{center}
\end{table}

\setlength{\tabcolsep}{8pt}	
\begin{table}[t]
	\footnotesize		
	\begin{center}
		\caption{Few-shot object detection performance on MS COCO. * means applying our w-dropout.}
		\label{table:21}
		\begin{tabular}{ccccccc}
			\hline\noalign{\smallskip}
			&\multicolumn{2}{c}{novel AP}&\multicolumn{2}{c}{novel AP50}&	\multicolumn{2}{c}{novel AP75}\\
			\hline\noalign{\smallskip}
			Method/Shot& 10 & 30 & 10 & 30 & 10 & 30 \\
			\noalign{\smallskip}
			\hline
			\noalign{\smallskip}
			Meta R-CNN~\cite{yan2019meta} &8.7&12.4&19.1&25.3&6.6&10.8\\	
			Meta R-CNN* &{\bf9.2}&{\bf12.7}&{\bf19.8}&{\bf25.9}&{\bf6.8}&{\bf11}\\
			\hline
			\noalign{\smallskip}
			TFA (Our Impl.)&8.9&12&16.2&21.2&8.9&12.3\\
			TFA*&{\bf9.7}&{\bf12.8}&{\bf17.5}&{\bf22.3}&{\bf9.8}&{\bf13.1}\\
			\hline				
		\end{tabular}
	\end{center}
\end{table}

\subsection{Implementation Details}
Meta R-CNN~\cite{yan2019meta}, TFA~\cite{wang2020frustratingly}, and FSCE~\cite{sun2021fsce} are adopted as our baselines. 
Faster R-CNN is used as the detector and ResNet-101 with a Feature Pyramid Network~\cite{lin2017feature} is the backbone. 
During experiments, we find using dropblock~\cite{ghiasi2018dropblock} to implement w-dropout is better than using normal dropout~\cite{srivastava2014dropout}. 
The training strategies of our method are the same as the selected baseline.

\subsection{Comparison with State-of-the-art Methods}

Based on Meta R-CNN, TFA, and FSCE, we apply w-dropout to get Meta R-CNN*, TFA* and FSCE* in Table.~\ref{table:1}. Specifically, Meta R-CNN* achieves more obvious improvement in almost all settings. Notably, Meta R-CNN* can gain 10$\%$ improvement in the setting of split 1 with 1-shot. TFA* is able to obtain higher accuracy than the baseline in all settings.
In general, our models get the highest improvement in the setting of 1-shot, followed by 3-shot, and the models get the least improvement in the 10-shot setting. With the decrease of the number of novel samples, $w$ in pre-training which presents the power of generalization becomes more important. Due to the few numbers of novel samples in the second training stage, we use multiple random seeds to get more stable results, as shown in Table.~\ref{table:1}. While comparing to recent methods, we use a certain seed to obtain the results, in order to be consistent with others. Our FSCE* can significantly outperform the baseline and achieve the state-of-the-art performance. 

Few-shot detection results of 10-shot and 30-shot setups for MS COCO are shown in Table.~\ref{table:21}.	
Our methods TFA* can achieve about 1$\%$ gain in most metrics. Meta R-CNN* also achieves better performance than the baseline.
It shows that the improvement of TFA and Meta R-CNN on MS COCO is lower than
that on PASCAL VOC, since MS COCO is a more challenging dataset.

\subsection{Ablation Study}
\subsubsection{W-dropout}
We follow most of the few-shot methods~\cite{yan2019meta,sun2021fsce} to do ablation studies with the dataset of VOC 1 split. 
We apply w-dropout on the pre-training stage and get the model as the initialization for fine-tuning. W-dropout is implemented by dropblock. W-dropout on Meta R-CNN can achieve 1$\%$ improvement on average presented in Table.~\ref{table:2}. Meanwhile, w-dropout can help TFA achieve higher performance (+4$\%$ on average) in all settings. In Table.~\ref{table:1}, w-dropout on FSCE brings 2$\%$-3$\%$ improvement on average. All comparison results show that our w-dropout can improve the baselines in a fair setting.

\begin{table}[t]
	\scriptsize
	\setlength{\tabcolsep}{4pt}		
	\begin{center}
		\caption{Results of novel mAP on the split 1 of VOC 2007 test set by using w-dropout on Meta R-CNN with batch size 1 and 4 (original setting) and TFA with batch size 16 (original setting). }
		\label{table:2}
		\begin{tabular}{ccccccc}
			\hline\noalign{\smallskip}
			&\multicolumn{3}{c}{Meta R-CNN}&\multicolumn{3}{c}{TFA}\\
			\hline\noalign{\smallskip}
			Setting / Shot (bs=1/4)& 1 & 3 & 10& 1 & 3 & 10\\
			\noalign{\smallskip}
			\hline
			\noalign{\smallskip}
			Baselines&23.3/14.7&{\bf37.8}/32.8&49.3/{\bf51.9}&22.4&40.3&53.1\\
			+ w-dropout&{\bf24.7}/{\bf16.6}&37.3/{\bf32.9}&{\bf51.8}/51.4&{\bf26.3}&{\bf45.6}&{\bf55.8}\\
			\hline
		\end{tabular}
	\end{center}
\end{table}

\begin{table}[t]
	\scriptsize
	\setlength{\tabcolsep}{6pt}		
	\begin{center}
		\caption{Results of novel mAP on the split 1 of VOC 2007 test set by adopting w-dropout (W) and dropout (D).}
		\label{table:4}
		\begin{tabular}{ccccccc}
			
			\hline\noalign{\smallskip}
			&\multicolumn{3}{c}{Meta R-CNN}&\multicolumn{3}{c}{TFA}\\
			\hline\noalign{\smallskip}
			Setting / Shot & 1 & 3 & 10&1&3&10\\
			\noalign{\smallskip}
			\hline
			\noalign{\smallskip}
			Baseline&14.7&32.8&{\bf51.9}&22.4&40.3&53.1\\
			+ W &{\bf16.2}&{\bf32.9}&51.4&{\bf25.1}&{\bf41.9}&53\\
			+ D &13.3&32.5&50&23.1&41.8&{\bf54.8}\\
			+ W$\&$D&14.5&32.5&48.8&25&41&50.5\\
			\hline
		\end{tabular}
	\end{center}
\end{table}

\begin{table}[b]
	\scriptsize
	\setlength{\tabcolsep}{3pt}	
	\begin{center}
		\caption{Results of base and novel mAP on the split 1\&2 of VOC 2007 test set by adopting w-dropout on TFA. }
		\label{table:100}
		\begin{tabular}{ccccccc}
			\hline\noalign{\smallskip}
			&\multicolumn{3}{c}{split 1} & 	\multicolumn{3}{c}{split2}\\
			\hline\noalign{\smallskip}
			Base / Novel & 1 & 3 &10 & 1 & 3&10 \\
			\hline
			\noalign{\smallskip}
			TFA&79.8/37.2&79.1/{\bf44.4}&{\bf79.4}/53.4&79.8/{\bf21.6}&78.6/34.8&78.3/38.4\\
			TFA*&{\bf80}/{\bf39.7}&{\bf79.7}/43.4&79.2/ {\bf55.2}&{\bf80.6}/20.3&{\bf78.8}/{\bf36.6}&{\bf78.7}/{\bf41.8}\\					
			\hline
		\end{tabular}
	\end{center}
\end{table}

\subsubsection{W-dropout vs Dropout}
W-dropout is applied on the transferable knowledge in the pre-training stage, while dropout is applied to the task-specific knowledge in the fine-tuning stage that is similar to the common one-stage training methods. W-dropout is for generalization power, normal dropout is for specific tasks. 
In order to compare w-dropout with dropout, we adopt w-dropout, dropout, and w-dropout$\&$dropout on Meta R-CNN and TFA. 
From the results in Table.~\ref{table:4}, only applying w-dropout obtains higher accuracy than using dropout or both strategies in most settings. 
The results of applying w-dropout in Table.~\ref{table:4} are different from that in Table.~\ref{table:2}. The reason is that w-dropout is implemented by our best setting (dropblock) in Table.~\ref{table:2}. In table.~\ref{table:4}, the normal dropout is used for showing the effectiveness of our idea in fair.

\begin{table}[t]
	\scriptsize
	\setlength{\tabcolsep}{6pt}		
	\begin{center}
		\caption{Results of mAP of novel classes on the split 1 of VOC 2007 test set by adopting different types of w-dropout on Meta R-CNN and TFA.}
		\label{table:81}
		\begin{tabular}{ccccccc}
			\hline\noalign{\smallskip}
			&\multicolumn{3}{c}{Meta R-CNN}&\multicolumn{3}{c}{TFA}\\
			\hline\noalign{\smallskip}
			Setting / Shot & 1 & 3 & 10&1&3&10\\
			\noalign{\smallskip}
			\hline
			\noalign{\smallskip}
			Baseline&14.7&32.8&{\bf51.9}&22.4&40.3&53.1\\
			+ dropout &14.3&31.3&50.2&25.1&41.9&53\\
			+ dropblock &{\bf16.2}&{\bf32.9}&51.4&{\bf26.7}&{\bf44}&{\bf54.2}\\
			\noalign{\smallskip}
            \hline
            \noalign{\smallskip}			
			on last conv layer &15.6&32.2&50.8&25.9&42.5&54.5\\
			on group 4 &{\bf16.2}&{\bf32.9}&{\bf51.4}&{\bf26.7}&44&54.2\\
			on group 3$\&$4 &16.1&32.5&50&26.3&{\bf45.6}&{\bf55.8}\\
			\hline
		\end{tabular}
	\end{center}
\end{table}

\subsubsection{Implementation of W-dropout}
W-dropout can be implemented by normal dropout or dropblock. 
In Table.~\ref{table:81}, the results show that using dropblock can get higher accuracy. Based on Meta R-CNN with dropblock, applying w-dropout on group 4 of the backbone is better than on group 3$\&$4 in. Using w-dropout on group 3$\&$4 of TFA gets the highest performance in most settings. 
Specifically, group 3 or 4 means the last convolution layer of each group.

\subsubsection{The Influence of W-dropout on Base and Novel Classes}
Based on TFA, we study the influence of w-dropout on the performance of base and novel classes. The results shown in Table.~\ref{table:100} prove that w-dropout improves the accuracy of novel classes without hurting the performance of base classes.

\section{Few-shot Classification}

\subsection{Datasets}
We follow the commonly used datasets~\cite{chen2019closer,yang2021free}, such as Caltech-UCSD Birds-200-2011 (CUB) and mini-ImageNet in few-shot classification. CUB is for fine-grained classification, which contains 200 classes. We follow the evaluation protocol of~\cite{hilliard2018few} that 200 classes are divided into 100 base, 50 validation, and 50 novel classes, respectively. The mini-ImageNet that consists of 100 categories is a subset of ImageNet, and each class contains 600 images of size 84$\times$84. Follow the way of splitting the dataset in~\cite{finn2017model,snell2017prototypical,sung2018learning}, the selected 100 classes are divided into 64 training classes, 16 validation classes, and 20 test classes. 


\begin{table}[t]
	\scriptsize
	\setlength{\tabcolsep}{1pt}		
	\begin{center}
		\caption{Few-shot classification results for both the mini-ImageNet and CUB datasets. }
		\label{table:13}
		\begin{tabular}{cccccc}
			\hline\noalign{\smallskip}
			&&\multicolumn{2}{c}{CUB} & 	\multicolumn{2}{c}{mini-ImageNet}\\
			\hline\noalign{\smallskip}
			Method&backbone& 1-shot & 5-shot & 1-shot & 5-shot \\
			\noalign{\smallskip}
			\hline
			\noalign{\smallskip}
			MatchingNet~\cite{vinyals2016matching}&Conv-4&60.52$\pm$0.88&75.29$\pm$0.75 &48.14$\pm$0.78 &63.48$\pm$0.66\\
			ProtoNet~\cite{snell2017prototypical}&Conv-4&50.46$\pm$0.88& 76.39$\pm$0.64& 44.42$\pm$0.84 &64.24$\pm$0.72\\
			MAML~\cite{finn2017model}&Conv-4&54.73$\pm$0.97 &75.75$\pm$0.76 &46.47$\pm$0.82& 62.71$\pm$0.71\\
			RelationNet~\cite{sung2018learning}&Conv-4&62.34$\pm$0.94 &77.84$\pm$0.68& 49.31$\pm$0.85& 66.60$\pm$0.69\\
			Baseline~\cite{chen2019closer}&Conv-4&47.12$\pm$0.74 &64.16$\pm$0.71 &42.11$\pm$0.71& 62.53$\pm$0.69\\
			Baseline++~\cite{chen2019closer}&Conv-4&60.53$\pm$0.83 &79.34$\pm$0.61&48.24$\pm$0.75& 66.43$\pm$0.63\\
			Baseline++ (Our Impl.)&Conv-4&60.95$\pm$0.87&78.38$\pm$0.63&47.60$\pm$0.73&65.74$\pm$0.64\\
			Baseline++*&Conv-4&{\bf63.61$\pm$0.92}&{\bf80.00$\pm$0.62}&{\bf50.88$\pm$0.74}&{\bf68.27$\pm$0.65}\\
			\hline
			\noalign{\smallskip}
			Baseline++(Our Impl.) &ResNet-10&63.27$\pm$0.97&80.24$\pm$0.59&53.36$\pm$0.79&73.85$\pm$0.64\\
			Baseline++*&ResNet-10&{\bf69.05$\pm$0.88}&{\bf82.92$\pm$0.54}&{\bf56.34$\pm$0.78}&{\bf75.58$\pm$0.59}\\	
			\hline
		\end{tabular}
	\end{center}
\end{table}

\subsection{Implementation Details}

We choose the Baseline++~\cite{chen2019closer}, which is a representative few-shot classification method, as our baseline. Based on this method, we can prove the effectiveness of our w-dropout in few-shot classification. During experiments, we follow the same training strategies in ~\cite{chen2019closer}. Specifically, Baseline++ is trained 200 epochs for the CUB dataset, and 400 epochs for the mini-ImageNet dataset. 
The evaluation setting we used is the same as the Baseline++. (600 randomly
episodes with the 95$\%$ confidence intervals) 

\subsection{Comparison with Baselines}
We report the accuracy of few-shot classification in Table.~\ref{table:13}, in which we build our Baseline++* by applying w-dropout on Baseline++. We use dropblock which is applied on the last convolution layer. 
The results demonstrate that our model outperforms others and using w-dropout can significantly improve the performance of Baseline++ in almost all settings.

\subsection{Ablation Study}

\subsubsection{W-dropout vs Dropout}
We apply w-dropout and dropout to show the importance of improving the generalization power of  transferable knowledge. Using normal dropout to implement our w-dropout for fair comparison.  The results are shown in Table.~\ref{table:15}. In the 5-shot setting of the CUB dataset, using w-dropout is comparable to the Baseline++. However, applying w-dropout can achieve the best performance in all other settings. 

\begin{table}[t]
	\scriptsize
	\setlength{\tabcolsep}{4pt}		
	\begin{center}
		\caption{Results of Baseline++ on the CUB and mini-ImageNet datasets by applying w-dropout (W) and dropout (D). }
		\label{table:15}
		\begin{tabular}{ccccc}
			\hline\noalign{\smallskip}
			&\multicolumn{2}{c}{CUB} & 	\multicolumn{2}{c}{mini-ImageNet}\\
			\hline\noalign{\smallskip}
			Method& 1-shot & 5-shot & 1-shot & 5-shot \\
			\noalign{\smallskip}
			\hline
			\noalign{\smallskip}
			Baseline++ (Our Impl.)&60.95$\pm$0.87&{\bf78.38$\pm$0.63}&47.60$\pm$0.73&65.74$\pm$0.64\\
			+ W$\&$D&58.34$\pm$0.88&75.76$\pm$0.64&44.67$\pm$0.68&63.80$\pm$0.66\\
			+ D&59.05$\pm$0.87&76.39$\pm$0.65&43.92$\pm$0.69&62.66$\pm$0.67\\
			+ W&{\bf62.71$\pm$0.87}&78.12$\pm$0.62&{\bf50.47$\pm$0.72}&{\bf68.20$\pm$0.65}\\		
			\hline
		\end{tabular}
	\end{center}
\end{table}

\section{Conclusion}
In this paper, we propose to use dropout to improve the generalization power of few-shot models. We conduct extensive experiments on the few-shot object detection and image classification tasks, and our models demonstrate great superiority towards the current few-shot methods. Moreover, our paper shows a new direction for few-shot models that is exploring the power of regularization techniques on few-shot models.


\bibliographystyle{IEEE}
\bibliography{strings,refs}

\begin{thebibliography}{10}

\bibitem{ren2016faster}
Shaoqing Ren, Kaiming He, Ross Girshick, and Jian Sun,
\newblock ``Faster r-cnn: Towards real-time object detection with region
  proposal networks,''
\newblock {\em IEEE transactions on pattern analysis and machine intelligence},
  vol. 39, no. 6, pp. 1137--1149, 2016.

\bibitem{dai2016r}
Jifeng Dai, Yi~Li, Kaiming He, and Jian Sun,
\newblock ``R-fcn: Object detection via region-based fully convolutional
  networks,''
\newblock {\em arXiv preprint arXiv:1605.06409}, 2016.

\bibitem{srivastava2014dropout}
Nitish Srivastava, Geoffrey Hinton, Alex Krizhevsky, Ilya Sutskever, and Ruslan
  Salakhutdinov,
\newblock ``Dropout: a simple way to prevent neural networks from
  overfitting,''
\newblock {\em The journal of machine learning research}, vol. 15, no. 1, pp.
  1929--1958, 2014.

\bibitem{everingham2010pascal}
Mark Everingham, Luc Van~Gool, Christopher~KI Williams, John Winn, and Andrew
  Zisserman,
\newblock ``The pascal visual object classes (voc) challenge,''
\newblock {\em International journal of computer vision}, vol. 88, no. 2, pp.
  303--338, 2010.

\bibitem{lin2014microsoft}
Tsung-Yi Lin, Michael Maire, Serge Belongie, James Hays, Pietro Perona, Deva
  Ramanan, Piotr Doll{\'a}r, and C~Lawrence Zitnick,
\newblock ``Microsoft coco: Common objects in context,''
\newblock in {\em European conference on computer vision}. Springer, 2014, pp.
  740--755.

\bibitem{yan2019meta}
Xiaopeng Yan, Ziliang Chen, Anni Xu, Xiaoxi Wang, Xiaodan Liang, and Liang Lin,
\newblock ``Meta r-cnn: Towards general solver for instance-level low-shot
  learning,''
\newblock in {\em Proceedings of the IEEE International Conference on Computer
  Vision}, 2019, pp. 9577--9586.

\bibitem{wang2020frustratingly}
Xin Wang, Thomas~E Huang, Trevor Darrell, Joseph~E Gonzalez, and Fisher Yu,
\newblock ``Frustratingly simple few-shot object detection,''
\newblock {\em arXiv preprint arXiv:2003.06957}, 2020.

\bibitem{sun2021fsce}
Bo~Sun, Banghuai Li, Shengcai Cai, Ye~Yuan, and Chi Zhang,
\newblock ``Fsce: Few-shot object detection via contrastive proposal
  encoding,''
\newblock in {\em Proceedings of the IEEE/CVF Conference on Computer Vision and
  Pattern Recognition}, 2021, pp. 7352--7362.

\bibitem{kang2019few}
Bingyi Kang, Zhuang Liu, Xin Wang, Fisher Yu, Jiashi Feng, and Trevor Darrell,
\newblock ``Few-shot object detection via feature reweighting,''
\newblock in {\em Proceedings of the IEEE International Conference on Computer
  Vision}, 2019, pp. 8420--8429.

\bibitem{li2021beyond}
Bohao Li, Boyu Yang, Chang Liu, Feng Liu, Rongrong Ji, and Qixiang Ye,
\newblock ``Beyond max-margin: Class margin equilibrium for few-shot object
  detection,''
\newblock in {\em Proceedings of the IEEE/CVF Conference on Computer Vision and
  Pattern Recognition}, 2021, pp. 7363--7372.

\bibitem{wu2020multi}
Jiaxi Wu, Songtao Liu, Di~Huang, and Yunhong Wang,
\newblock ``Multi-scale positive sample refinement for few-shot object
  detection,''
\newblock in {\em European Conference on Computer Vision}. Springer, 2020, pp.
  456--472.

\bibitem{li2021few}
Yiting Li, Haiyue Zhu, Yu~Cheng, Wenxin Wang, Chek~Sing Teo, Cheng Xiang,
  Prahlad Vadakkepat, and Tong~Heng Lee,
\newblock ``Few-shot object detection via classification refinement and
  distractor retreatment,''
\newblock in {\em Proceedings of the IEEE/CVF Conference on Computer Vision and
  Pattern Recognition}, 2021, pp. 15395--15403.

\bibitem{zhang2021hallucination}
Weilin Zhang and Yu-Xiong Wang,
\newblock ``Hallucination improves few-shot object detection,''
\newblock in {\em Proceedings of the IEEE/CVF Conference on Computer Vision and
  Pattern Recognition}, 2021, pp. 13008--13017.

\bibitem{lin2017feature}
Tsung-Yi Lin, Piotr Doll{\'a}r, Ross Girshick, Kaiming He, Bharath Hariharan,
  and Serge Belongie,
\newblock ``Feature pyramid networks for object detection,''
\newblock in {\em Proceedings of the IEEE conference on computer vision and
  pattern recognition}, 2017, pp. 2117--2125.

\bibitem{ghiasi2018dropblock}
Golnaz Ghiasi, Tsung-Yi Lin, and Quoc~V Le,
\newblock ``Dropblock: A regularization method for convolutional networks,''
\newblock {\em arXiv preprint arXiv:1810.12890}, 2018.

\bibitem{chen2019closer}
Wei-Yu Chen, Yen-Cheng Liu, Zsolt Kira, Yu-Chiang~Frank Wang, and Jia-Bin
  Huang,
\newblock ``A closer look at few-shot classification,''
\newblock {\em arXiv preprint arXiv:1904.04232}, 2019.

\bibitem{yang2021free}
Shuo Yang, Lu~Liu, and Min Xu,
\newblock ``Free lunch for few-shot learning: Distribution calibration,''
\newblock {\em arXiv preprint arXiv:2101.06395}, 2021.

\bibitem{hilliard2018few}
Nathan Hilliard, Lawrence Phillips, Scott Howland, Art{\"e}m Yankov, Courtney~D
  Corley, and Nathan~O Hodas,
\newblock ``Few-shot learning with metric-agnostic conditional embeddings,''
\newblock {\em arXiv preprint arXiv:1802.04376}, 2018.

\bibitem{finn2017model}
Chelsea Finn, Pieter Abbeel, and Sergey Levine,
\newblock ``Model-agnostic meta-learning for fast adaptation of deep
  networks,''
\newblock in {\em International Conference on Machine Learning}. PMLR, 2017,
  pp. 1126--1135.

\bibitem{snell2017prototypical}
Jake Snell, Kevin Swersky, and Richard Zemel,
\newblock ``Prototypical networks for few-shot learning,''
\newblock in {\em Advances in neural information processing systems}, 2017, pp.
  4077--4087.

\bibitem{sung2018learning}
Flood Sung, Yongxin Yang, Li~Zhang, Tao Xiang, Philip~HS Torr, and Timothy~M
  Hospedales,
\newblock ``Learning to compare: Relation network for few-shot learning,''
\newblock in {\em Proceedings of the IEEE Conference on Computer Vision and
  Pattern Recognition}, 2018, pp. 1199--1208.

\bibitem{vinyals2016matching}
Oriol Vinyals, Charles Blundell, Timothy Lillicrap, Daan Wierstra, et~al.,
\newblock ``Matching networks for one shot learning,''
\newblock {\em Advances in neural information processing systems}, vol. 29, pp.
  3630--3638, 2016.

\end{thebibliography}

\end{document}